\pdfoutput=1 % arXiv: compile with pdfLaTeX (PNG/PDF figures)
\documentclass[pdflatex,sn-mathphys-num]{sn-jnl}% Math and Physical Sciences Numbered Reference Style

\usepackage{graphicx}%
\usepackage{multirow}%
\usepackage{amsmath,amssymb,amsfonts}%
\usepackage{amsthm}%
\usepackage{mathrsfs}%
\usepackage[title]{appendix}%
\usepackage{xcolor}%
\usepackage{textcomp}%
\usepackage{manyfoot}%
\usepackage{booktabs}%
\usepackage{algorithm}%
\usepackage{algorithmic}%
\usepackage{subfig}%
\usepackage{float}%
\usepackage{bm}%
\usepackage{url}%
\theoremstyle{thmstyleone}%
\theoremstyle{thmstyletwo}%
\theoremstyle{thmstylethree}%

\usepackage{etoolbox}%
\makeatletter
\patchcmd{\@maketitle}{{\ifnum\aucount>1*\fi}Corresponding author(s). E-mail(s): }{}{}{}
\patchcmd{\@maketitle}{Contributing authors:\ }{}{}{}
\patchcmd{\@maketitle}{\corrauthemail\par}{\corrauthemail}{}{}
\patchcmd{\@maketitle}{\ifequalcont}%
  {\footnotetext[1]{Corresponding author: Mohammed El Hanjri
   (\texttt{mohammed.elhanjri@um5r.ac.ma}).}\ifequalcont}{}{}
\makeatother

\begin{document}

\title[Opinion Dynamics-based Coalition Formation for FL in Heterogeneous IoT Systems]{Opinion Dynamics-based Coalition Formation for Federated Learning in Heterogeneous IoT Systems}

\author*[1]{\fnm{Mohammed} \sur{El Hanjri}}\email{mohammed.elhanjri@um5r.ac.ma}

\author[1]{\fnm{Anas} \sur{Abouaomar}}\email{anas\_abouaomar@um5.ac.ma}

\author[2]{\fnm{Hamidou} \sur{Tembine}}\email{hamidou.tembine@uqtr.ca}

\author[1]{\fnm{Abdellatif} \sur{Kobbane}}\email{abdellatif.kobbane@ensias.um5.ac.ma}

\affil[1]{\orgname{ENSIAS, Mohammed V University in Rabat}, \orgaddress{\country{Morocco}}}

\affil[2]{Department of Electrical and Computer Engineering, School of Engineering, University of Quebec at Trois-Rivieres, Quebec, Canada}

\abstract{Federated learning (FL) enables privacy-preserving, on-device training
across heterogeneous Internet-of-Things (IoT) deployments such as smart-city
water-metering networks, where each smart meter observes a household-specific
consumption time series. Under such statistical heterogeneity, the standard
Federated Averaging (FedAvg) aggregation averages dissimilar local models into
a single global model that may fail to capture client-specific patterns. We
address this by forming
client coalitions directly in the local-weight space and aggregating at the
coalition level. Extending a prior weight-driven coalition-formation scheme, we
model coalition formation as a Hegselmann--Krause (HK) bounded-confidence
opinion-dynamics process acting on the local weights, and develop variants of
the HK interaction based on Euclidean-distance and cosine-similarity confidence
criteria. The framework is applied to short-term water-consumption
forecasting with local Long Short-Term Memory (LSTM) models and evaluated
against FedAvg, Per-FedAvg, FedProx, and FedAvg with Euclidean-distance or
cosine-similarity coalition formation. Experiments on a real smart-metering dataset of water consumption show that the proposed HK-based coalition formation
produces stable, endogenous coalition structures within at most ten inner
iterations, incurs no additional client-side computation or communication compared
to FedAvg, and reduces the average MAE by up to 54\% relative to FedAvg,
39\% relative to FedProx, and 24\% relative to Per-FedAvg, while achieving
the highest global accuracy (83--85\%).}


\maketitle

\section{Introduction}
\label{sec:introduction}

Federated learning (FL) has become a standard paradigm for training models
collaboratively across large populations of distributed, privacy-sensitive
Internet-of-Things (IoT) devices~\cite{mcmahan2017communication,kairouz2021advances}.
By keeping raw data on each device and exchanging only model parameters, FL
preserves privacy and reduces communication relative to centralized training,
which makes it attractive for smart-city and edge
deployments~\cite{pandya2023federated,jiang2020federated,singh2022framework}. Its
effectiveness in practice, however, is limited by statistical
heterogeneity: IoT devices observe data from distinct, non-identical
distributions, so their local objectives conflict and the standard Federated
Averaging (FedAvg) aggregation blends dissimilar local models into a single global
model that is a poor compromise for individual devices~\cite{li2020federated}.
Improving the efficiency of FL under such heterogeneity is the problem this paper
addresses.\\

We study this problem through a representative heterogeneous IoT use case:
short-term load forecasting (STLF) of water consumption in smart cities. Networked
smart meters record household-level consumption that is highly heterogeneous,
shaped by behavioral, demographic, and environmental factors that differ widely
across households~\cite{kavya2023short,drogkoula2023comprehensive}, and the data
are privacy-sensitive, since fine-grained traces can reveal occupancy and activity
patterns. Each device trains a local Long Short-Term Memory (LSTM) forecaster,
well suited to such series through its gated memory of long-range temporal
dependencies~\cite{nasser2020two,bezzar2022data}. This setting exhibits exactly
the non-IID structure that challenges FL and serves as the testbed for our
experiments; the proposed method itself is independent of the forecasting model
and operates only on the exchanged weights.\\

A natural response to heterogeneity is to aggregate not across all devices at
once, but within groups of compatible clients whose models can be combined
without mutual interference; the advantage of such personalized or group-wise
models over a single shared model is well documented for load
forecasting~\cite{reguieg2023comparative,xu2024decentralized}. The
key question is how to form these groups without access to the private data: a
privacy-compatible signal is the local model itself, since clients with similar
data tend to learn similar weights. This idea has been exploited through
weight-driven coalition formation with a fixed number of distance-based
coalitions~\cite{el2024efficient}.\\

In this paper we generalize that scheme by casting coalition formation as a
bounded-confidence opinion-dynamics process on the local weights.
Specifically, we apply the Hegselmann--Krause (HK) model~\cite{rainer2002opinion}
to the post-training local models: each client is an agent whose ``opinion'' is
its weight vector, and agents reinforce one another only when their models are
mutually compatible. Running this interaction to its fixed points yields an
endogenous partition, since the number and membership of coalitions emerge
from the data rather than being fixed in advance, and it naturally isolates atypical
clients as outliers instead of forcing them into a coalition. We study different
notions of compatibility: a Euclidean confidence ball, a cosine-similarity
threshold, and an asymmetric cosine-confidence bound, motivated by the observation
that the direction of high-dimensional weights is often more informative than
their absolute distance. In this paper the contributions are as follows.
\begin{itemize}
    \item We formulate heterogeneity-aware FL aggregation as a
    coalition-formation problem on the local weights, and show how a
    HK bounded-confidence interaction yields coalitions whose
    number and membership are endogenous and adaptive.
    \item We develop three instantiations of the interaction (Euclidean,
    cosine-similarity, and asymmetric cosine-confidence) and a coalition
    aggregation algorithm that combines coalition barycenters into the shared
    model, with a complexity analysis showing no additional client-side computation or
    communication relative to FedAvg.
    \item We instantiate the framework for short-term water-consumption
    forecasting with local LSTM models and evaluate it on a smart-city dataset
    against FedAvg with Euclidean coalitions, FedAvg with cosine coalitions,
    FedProx~\cite{li2020federated}, and Per-FedAvg~\cite{reguieg2023comparative},
    reporting coalition structure, convergence, and forecasting accuracy.
\end{itemize}

The remainder of the paper is organized as follows. Section~\ref{sec:related} reviews related work. Section~\ref{sec:system_model} presents the system model and the problem formulation. Section~\ref{sec:framework} develops the proposed HK-based coalition-formation
framework. Sections~\ref{sec:setup} and~\ref{sec:results} describe the experimental setup and discuss the results, and the conclusion 
and percpectives in Section~\ref{sec:conclusion}.

\section{Related Work}
\label{sec:related}

%\subsection{Federated learning under statistical heterogeneity}
%\label{ssec:rw_fl}

FL was introduced as a communication-efficient, privacy-preserving
alternative to centralized training~\cite{mcmahan2017communication,hard2018federated},
and surveys document its breadth and open
problems~\cite{kairouz2021advances}, including smart-city
applications~\cite{pandya2023federated,jiang2020federated} and privacy-preserving
IoT deployments~\cite{singh2022framework}. A central obstacle is statistical
heterogeneity: when client data are non-IID, FedAvg converges slowly and the
resulting global model may deviate substantially from individual client
optima. FedProx~\cite{li2020federated} adds a proximal
term to the local objective to limit client drift, improving stability under
heterogeneity. Federated approaches have also been applied to energy and demand
management at the edge~\cite{rezazadeh2022federated,el2026federated}, with
incentive mechanisms studied to encourage participation~\cite{khan2020federated}.
Our work targets the same heterogeneity obstacle but, rather than regularizing a
single global model, restructures the aggregation itself around coalitions of
compatible clients.\\

%\subsection{Clustered and personalized FL, and coalition formation}
%\label{ssec:rw_cluster}

A complementary line of work produces multiple or client-specific models.
Personalized FL adapts the shared model to each client, for instance the
meta-learning-based Per-FedAvg, evaluated against FedAvg under Dirichlet-skewed
data in~\cite{reguieg2023comparative}, whereas decentralized, heterogeneity-aware
schemes have been proposed for smart-home energy and comfort
prediction~\cite{xu2024decentralized}. Clustering ideas, classical in edge and
sensor networks (e.g.\ $k$-means grouping of devices~\cite{el2016energy}), have
been transposed to FL to aggregate within groups of similar clients. Closest to
this paper, the scheme of~\cite{el2024efficient} forms coalitions
directly in the local-weight space using a distance between models and a fixed
number of coalitions. The present paper differs in that the coalitions are not
prescribed: they emerge as fixed points of a bounded-confidence interaction, and
we additionally introduce cosine-based compatibility, which is invariant to weight
magnitude.\\

Beyond fixed-size coalitions, a growing body of recent work clusters FL
clients directly from their model parameters. WSCC~\cite{tian2022wscc}
applies affinity propagation to the cosine distance between client weight
vectors, so that the number of clusters is determined automatically rather
than prescribed; FedClust~\cite{islam2024fedclust} forms clusters in a
one-shot manner from strategically selected partial weights of the locally
trained models, avoiding both the long stabilization phase of iterative
clustered FL and a predefined number of groups. Game-theoretic formulations
instead treat grouping as an explicit coalition game:
DualGFL~\cite{chen2025dualgfl} couples a lower-level hedonic game, in which
clients form coalitions according to preference profiles, with an
upper-level auction in which coalitions bid for participation in training.
These works confirm that weight similarity is a reliable,
privacy-compatible proxy for data similarity, but they rely on generic
clustering heuristics or hand-designed utility functions. In contrast, we
obtain the partition as the set of fixed points of a bounded-confidence
interaction, a class of dynamics whose convergence and clustering
properties are supported by a mature theory~\cite{bernardo2024bounded},
which yields coalitions that are simultaneously endogenous, outlier-aware,
and free of any auxiliary clustering machinery.\\
%\subsection{Opinion dynamics and bounded-confidence models}
%\label{ssec:rw_opinion}

Bounded-confidence models from opinion dynamics describe how agents update their
positions by averaging only over peers within a confidence range. The
HK model~\cite{rainer2002opinion} is the canonical such
model: agents repeatedly move to the mean of the opinions lying within a
confidence bound, and the dynamics settle into clusters of consensus separated by
gaps. This clustering-by-confidence behavior is a natural fit for coalition
formation in FL, where each client holds a model (an opinion) and only compatible
models should be aggregated together. We are, to our knowledge, applying the HK
interaction to the local weights of an FL system to drive coalition formation,
and we extend it from the classical Euclidean ball to cosine-similarity and
asymmetric cosine-confidence neighborhoods suited to high-dimensional model
parameters.\\

The use case in which we evaluate the framework, short-term load and demand
forecasting, has its own substantial literature. Machine-learning models, and deep
recurrent networks in particular, are now standard for this task.
Marino et al.~\cite{marino2016building} used sequence-to-sequence LSTM
architectures for building energy load, with strong results at fine temporal
resolution but limited gains at coarser (hourly) granularity. Evolutionary and
feature-selection strategies have been used to tune LSTM
forecasters~\cite{almalaq2018evolutionary,bouktif2018optimal}, though their
accuracy often degrades when transferred to new datasets. To cope with consumption
variability, some works group users with similar profiles or pool heterogeneous
series to broaden diversity~\cite{stephen2015incorporating,shi2017deep}. In the
water domain, LSTM- and micro-service-based systems have been proposed for urban
demand prediction~\cite{nasser2020two,kavya2023short,drogkoula2023comprehensive},
and time-series analysis has been applied to household-level
consumption~\cite{bezzar2022data}. These forecasting approaches are predominantly
centralized; the privacy and scalability constraints they face are precisely what
motivate the federated, heterogeneity-aware formulation studied here.\\

Federated formulations of short-term load forecasting itself have emerged
very recently. In \cite{rahman2025electrical}, the authors personalize
FL-based load prediction under non-IID metering data by adapting per-client
learning rates through meta-learning, and a follow-up work extends
federated forecasting to multihop smart-metering networks with limited
connectivity~\cite{rahman2025multihop}. Bose and
Kim~\cite{bose2023federated} instead keep selected personalization layers
of the forecasting model local to each client, so that shared layers
capture common temporal structure while private layers absorb heterogeneity
across buildings. These approaches validate FL as a practical substrate for
consumption forecasting, but they personalize each client individually
rather than restructuring the aggregation itself, and they focus on
electricity; group-level aggregation for water-consumption forecasting, as
studied here, remains largely unexplored.\\
%\subsection{Positioning}
%\label{ssec:rw_positioning}

In summary, FL restores privacy but struggles with statistical
heterogeneity, and existing remedies either regularize a single global model
(FedProx), personalize per client (Per-FedAvg), or cluster clients with a fixed
group structure (weight-driven coalitions~\cite{el2024efficient}); deep-learning forecasters,
meanwhile, are accurate but conventionally centralized. This paper unifies the
coalition idea with bounded-confidence opinion dynamics, yielding an endogenous,
outlier-aware partition of clients and a family of compatibility geometries, and
evaluates it on smart-city water-consumption forecasting as a representative
heterogeneous IoT use case.

\section{System Model and Problem Formulation}
\label{sec:system_model}

\begin{figure}[!htbp]
    \centering
    \includegraphics[width=\linewidth]{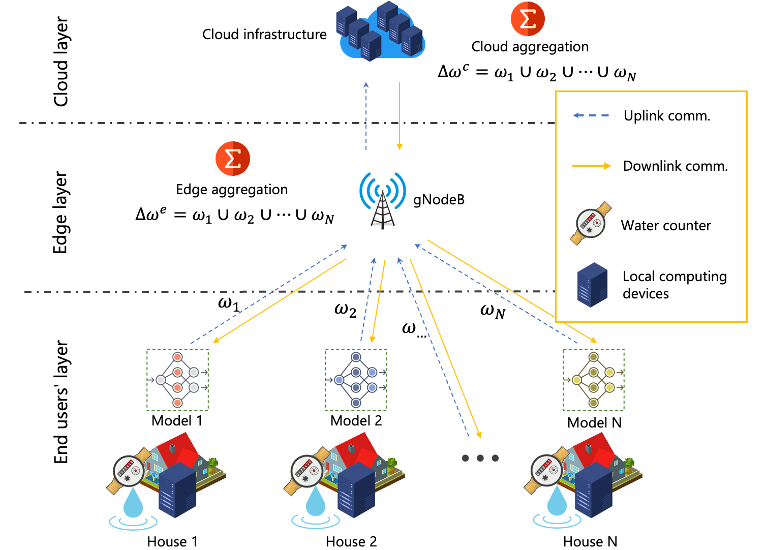}
    \caption{Network architecture: a single edge (MEC) server orchestrates
    federated training over $N$ IoT clients (smart meters); each client trains a
    local model on its private data and exchanges only model parameters.}
    \label{fig:sm1}
\end{figure}

We consider an FL system over a heterogeneous IoT network,
instantiated on a smart-city water-metering scenario. A set of edge clients
collaboratively trains forecasting models under the coordination of an edge
server, without sharing raw data. We keep the local learner generic and treat
the LSTM forecaster and the water-consumption data as the concrete instantiation
used in the experiments (Section~\ref{sec:setup}); the focus here is the FL
protocol and the statistical heterogeneity that makes a single shared model
inadequate. The overall architecture is shown in Figure~\ref{fig:sm1}, and the
notation is summarized in Table~\ref{tab:notation}.

\begin{table}[!htbp]
\centering
\caption{Summary of notation.}
\label{tab:notation}
\begin{tabular}{ll}
\toprule
\textbf{Symbol} & \textbf{Description} \\
\midrule
$\mathcal{N},\,N$ & set / number of IoT clients \\
$\mathcal{S}^{t},\,S$ & active subset in round $t$ / its size \\
$T,\,E,\,\eta$ & rounds / local epochs / learning rate \\
$\mathcal{D}_i,\,m_i,\,p_i$ & local dataset / size / weight $m_i/M$ \\
$\mathcal{P}_i,\,F_i,\,F$ & local distribution / local risk / global objective \\
$f(\cdot;\bm{\omega}),\,\bm{\omega}\in\mathbb{R}^{d}$ & local model (LSTM) / its parameters \\
$\bm{\omega}_i^{t},\,\bm{\theta}^{t}$ & local model of $i$ / shared model, round $t$ \\
$\zeta,\,\beta$ & heterogeneity constants, Eq.~\eqref{eq:bounded_dissimilarity} \\
$\Pi^{t},\,\mathcal{C}_k^{t},\,K^{t}$ & coalition partition / coalition $k$ / number of coalitions \\
$\mathbf{b}_k^{t}$ & barycenter of coalition $\mathcal{C}_k^{t}$ \\
$\bm{z}_i^{(s)}$ & HK opinion of client $i$ at inner step $s$ \\
$\varepsilon,\,\tau,\,(\varepsilon_\ell,\varepsilon_r)$ & Euclidean / cosine / asymmetric confidence parameters \\
\bottomrule
\end{tabular}
\end{table}
\subsection{System architecture}
\label{ssec:sm_protocol}

Let $\mathcal{N}=\{1,\dots,N\}$ be the set of clients, each an IoT device (in the
case study, a household smart meter) holding a private dataset and training
locally. A single MEC server~\cite{filali2020multi} orchestrates and aggregates;
it holds no data. Training runs over $T$ rounds: at round $t$ the server selects
an active subset $\mathcal{S}^{t}\subseteq\mathcal{N}$ with
$|\mathcal{S}^{t}|=S\le N$ (reflecting connectivity and energy constraints), and
only those clients compute and communicate.

Each client $i$ trains a model $f(\cdot;\bm{\omega})$ with parameters
$\bm{\omega}\in\mathbb{R}^{d}$; all clients share the same architecture, so the
parameter vectors are directly comparable. In round $t$, client $i$ starts from
the shared model $\bm{\theta}^{t-1}$ and runs $E$ local SGD epochs (learning rate
$\eta$) on its empirical risk
$F_i(\bm{\omega})=\frac{1}{m_i}\sum_{n=1}^{m_i}\ell(f(\mathbf{x}_i^{(n)};
\bm{\omega}),y_i^{(n)})$ with squared-error loss $\ell$,
\begin{equation}
    \bm{\omega}_i^{t,e+1}
    = \bm{\omega}_i^{t,e} - \eta\,\nabla F_i\!\bigl(\bm{\omega}_i^{t,e}\bigr),
    \quad e=0,\dots,E-1,\quad \bm{\omega}_i^{t,0}=\bm{\theta}^{t-1},
    \label{eq:local_sgd}
\end{equation}
producing the local model
$\bm{\omega}_i^{t}\triangleq\bm{\omega}_i^{t,E}$. It uploads only
$\bm{\omega}_i^{t}$, and the server
aggregates. Under FedAvg~\cite{mcmahan2017communication},
\begin{equation}
    \bm{\theta}^{t}
    = \sum_{i\in\mathcal{S}^{t}} \frac{m_i}{M^{t}}\,\bm{\omega}_i^{t},
    \qquad M^{t}=\sum_{i\in\mathcal{S}^{t}} m_i .
    \label{eq:fedavg}
\end{equation}
Equation~\eqref{eq:fedavg} is the baseline our coalition-aware aggregation
replaces. Only model parameters ever leave a device, and the server keeps updates
only transiently for aggregation, which preserves data locality under an
honest-but-curious server.

\subsection{Local data and learning task}
\label{ssec:sm_data}

Client $i$ holds a private dataset
$\mathcal{D}_i=\{(\mathbf{x}_i^{(n)},y_i^{(n)})\}_{n=1}^{m_i}$, $m_i=|\mathcal{D}_i|$,
drawn from a client-specific distribution $\mathcal{P}_i$, with
$M=\sum_{i}m_i$. In the case study the data are univariate water-consumption
series mapped to fixed-length look-back windows $\mathbf{x}_i^{(n)}$ with
one-step-ahead targets $y_i^{(n)}$ (windowing details in
Section~\ref{sec:setup}). The forecaster $f$ is an LSTM, suited to consumption
data through its gated memory of temporal
dependencies~\cite{nasser2020two,bezzar2022data}; the proposed framework is
agnostic to this choice and operates solely on the parameter vectors
$\bm{\omega}$.

\subsection{Statistical heterogeneity}
\label{ssec:sm_hetero}

Statistical heterogeneity is the central difficulty addressed here. Clients are
non-IID ($\mathcal{P}_i\neq\mathcal{P}_j$ for $i\neq j$), combining
distributional heterogeneity (distinct data-generating processes, so the
local minimizers
$\bm{\omega}_i^{\star}=\arg\min_{\bm{\omega}}\mathcal{F}_i(\bm{\omega})$ of the
population risks $\mathcal{F}_i$ differ across clients) and quantity skew
through unequal $m_i$. We quantify distributional heterogeneity by a standard
bounded-dissimilarity condition: there exist $\zeta\ge0$, $\beta\ge1$ with
\begin{equation}
    \frac{1}{N}\sum_{i\in\mathcal{N}}
    \bigl\|\nabla F_i(\bm{\omega})-\nabla F(\bm{\omega})\bigr\|_2^{2}
    \;\le\; \zeta^{2}+\beta^{2}\,\bigl\|\nabla F(\bm{\omega})\bigr\|_2^{2},
    \qquad\forall\bm{\omega},
    \label{eq:bounded_dissimilarity}
\end{equation}
where $\zeta=0$ is the IID case and large $\zeta$ signals conflicting local
optima. The global objective is the sample-weighted aggregate
\begin{equation}
    \min_{\bm{\omega}\in\mathbb{R}^{d}}\;
    F(\bm{\omega})=\sum_{i\in\mathcal{N}} p_i\,\mathcal{F}_i(\bm{\omega}),
    \qquad p_i=\frac{m_i}{M},
    \label{eq:global_objective}
\end{equation}
which FedAvg targets through \eqref{eq:fedavg}. The premise of this work is that
under large $\zeta$ a single minimizer of \eqref{eq:global_objective} is a poor
compromise, which motivates solving it at the level of coalitions of compatible
clients (Sections~\ref{sec:system_model}--\ref{sec:framework}). Beyond
\eqref{eq:bounded_dissimilarity}, the later discussion assumes only standard
conditions ($\rho$-smooth $F_i$, unbiased bounded-variance stochastic gradients,
a common parameterization); we make no convergence claim beyond what these
support.\\

% Building on the system model, we now make precise why FedAvg degrades under
% heterogeneity, why the geometry of the local
% weights is an informative and privacy-compatible signal of that heterogeneity
% (Section~\ref{ssec:pf_similarity}), and how this leads to a coalition-formation
% problem (Section~\ref{ssec:pf_statement}).

FedAvg returns a single model $\bm{\theta}$ that minimizes the aggregate
objective \eqref{eq:global_objective}. When clients are heterogeneous, the local
minimizers $\bm{\omega}_i^{\star}$ are spread out and the aggregate minimizer is
a compromise that may be far from each of them: for any $\bm{\theta}$,
\begin{equation}
    \sum_{i\in\mathcal{N}} p_i\,
    \bigl\|\bm{\omega}_i^{\star}-\bm{\theta}\bigr\|_2^{2}
    \;\ge\;
    \sum_{i\in\mathcal{N}} p_i\,
    \bigl\|\bm{\omega}_i^{\star}-\bar{\bm{\omega}}^{\star}\bigr\|_2^{2}
    \;=\; \mathrm{Var}_p\!\bigl(\bm{\omega}^{\star}\bigr),
    \qquad
    \bar{\bm{\omega}}^{\star}=\sum_{i} p_i\,\bm{\omega}_i^{\star},
    \label{eq:var_lb}
\end{equation}
i.e.\ no single model can be closer to all local optima than their weighted
dispersion. This dispersion grows with the heterogeneity constant $\zeta$ in
\eqref{eq:bounded_dissimilarity}: larger $\zeta$ implies more conflicting local
gradients and hence a larger $\mathrm{Var}_p(\bm{\omega}^{\star})$. Two further
effects compound this. First, with partial participation the per-round average in
\eqref{eq:fedavg} is taken over a sampled subset $\mathcal{S}^{t}$, so its
expectation matches \eqref{eq:global_objective} but its variance is inflated by
heterogeneity, slowing and destabilizing convergence. Second, a few strongly
atypical clients (outliers violating \eqref{eq:bounded_dissimilarity}) can drag
$\bm{\theta}$ away from the bulk of clients. A single global model is therefore a
structurally poor target when $\zeta$ is large; some form of client grouping is
needed so that aggregation is performed only among compatible clients.

\subsection{Local-weight similarity metrics }
\label{ssec:pf_similarity}

Grouping clients requires a measure of how similar their learning problems are.
The data distributions $\mathcal{P}_i$ are not observable at the server under the
privacy model of Section~\ref{ssec:sm_protocol}, but the post-training local
models $\bm{\omega}_i^{t}$ are exactly the quantities already exchanged. Because
each $\bm{\omega}_i^{t}$ results from local optimization toward
$\bm{\omega}_i^{\star}$, clients with similar distributions tend to produce
nearby weights, while dissimilar clients diverge in parameter space. The local
weights are thus a natural, privacy-compatible proxy for distributional
similarity, requiring no additional disclosure. We compare two geometries on
$\mathbb{R}^{d}$: the Euclidean distance
\begin{equation}
    d_E(\bm{\omega}_i^{t},\bm{\omega}_j^{t})
    = \bigl\|\bm{\omega}_i^{t}-\bm{\omega}_j^{t}\bigr\|_2,
    \label{eq:deuclid}
\end{equation}
which captures absolute deviation, and the cosine similarity
\begin{equation}
    \mathrm{cosim}(\bm{\omega}_i^{t},\bm{\omega}_j^{t})
    = \frac{\langle\bm{\omega}_i^{t},\bm{\omega}_j^{t}\rangle}
           {\|\bm{\omega}_i^{t}\|_2\,\|\bm{\omega}_j^{t}\|_2},
    \label{eq:cosim}
\end{equation}
which captures directional (angular) agreement and is invariant to the magnitude
of the weight vectors. The two need not agree: high-dimensional neural weights
can differ substantially in norm (e.g.\ because of quantity skew or unequal
local progress) while pointing in similar directions. In such cases cosine
similarity may identify clients with aligned learning dynamics that Euclidean
distance would separate, which motivates studying both geometries within the same
framework (Section~\ref{sec:framework}).

\subsection{Coalition-formation problem statement}
\label{ssec:pf_statement}

We replace the single-centroid aggregation of FedAvg by aggregation over a
partition of the active clients into coalitions of similar local models. At round
$t$, let
$\Pi^{t}=\{\mathcal{C}_1^{t},\dots,\mathcal{C}_K^{t}\}$ be a partition of
$\mathcal{S}^{t}$ into $K$ disjoint coalitions, and let
$\mathbf{b}_k^{t}=\frac{1}{|\mathcal{C}_k^{t}|}\sum_{i\in\mathcal{C}_k^{t}}
\bm{\omega}_i^{t}$ denote the barycenter of coalition $\mathcal{C}_k^{t}$. The
coalition-formation objective is to choose $\Pi^{t}$ so that within-coalition
dissimilarity is small, e.g.
\begin{equation}
    \min_{\Pi^{t}}\;
    \sum_{k=1}^{K}\sum_{i\in\mathcal{C}_k^{t}}
    d\!\bigl(\bm{\omega}_i^{t},\mathbf{b}_k^{t}\bigr),
    \label{eq:coalition_obj}
\end{equation}
where $d(\cdot,\cdot)$ is the chosen dissimilarity, derived from
\eqref{eq:deuclid} or \eqref{eq:cosim}. Aggregation is then carried out at the
coalition level rather than over all clients at once, so that dissimilar clients
no longer average into one another. Directly optimizing \eqref{eq:coalition_obj}
is a combinatorial clustering problem, and a fixed $K$ with hard assignment does
not adapt the number or membership of coalitions to the evolving weights or to
outliers. This motivates the bounded-confidence formulation of
Section~\ref{sec:framework}, in which coalitions emerge from a
HK interaction on the local weights: each client aggregates only
with peers inside a confidence neighborhood, so the partition forms endogenously
and atypical clients remain isolated rather than forced into a coalition.

\section{Proposed HK-based Coalition Formation Framework}
\label{sec:framework}

We now address the coalition-formation problem \eqref{eq:coalition_obj} by
letting coalitions emerge from a bounded-confidence interaction on the
local weights, rather than fixing their number or membership in advance. We first
state the general mechanism and its rationale
(Section~\ref{ssec:fw_hk}), then instantiate it with a Euclidean confidence ball
(Section~\ref{ssec:fw_euclid}), a cosine-similarity threshold
(Section~\ref{ssec:fw_cosine}), and an asymmetric cosine-confidence bound
(Section~\ref{ssec:fw_asym}), and finally give the coalition aggregation
algorithm (Section~\ref{ssec:fw_algo}). This extends the weight-driven coalition
scheme of~\cite{el2024efficient} by replacing its fixed-$K$ assignment with an
opinion-dynamics process whose fixed points define the coalitions.

\subsection{Coalitions as a bounded-confidence opinion dynamics}
\label{ssec:fw_hk}

\begin{figure*}[t]
    \centering
    \includegraphics[width=\linewidth]{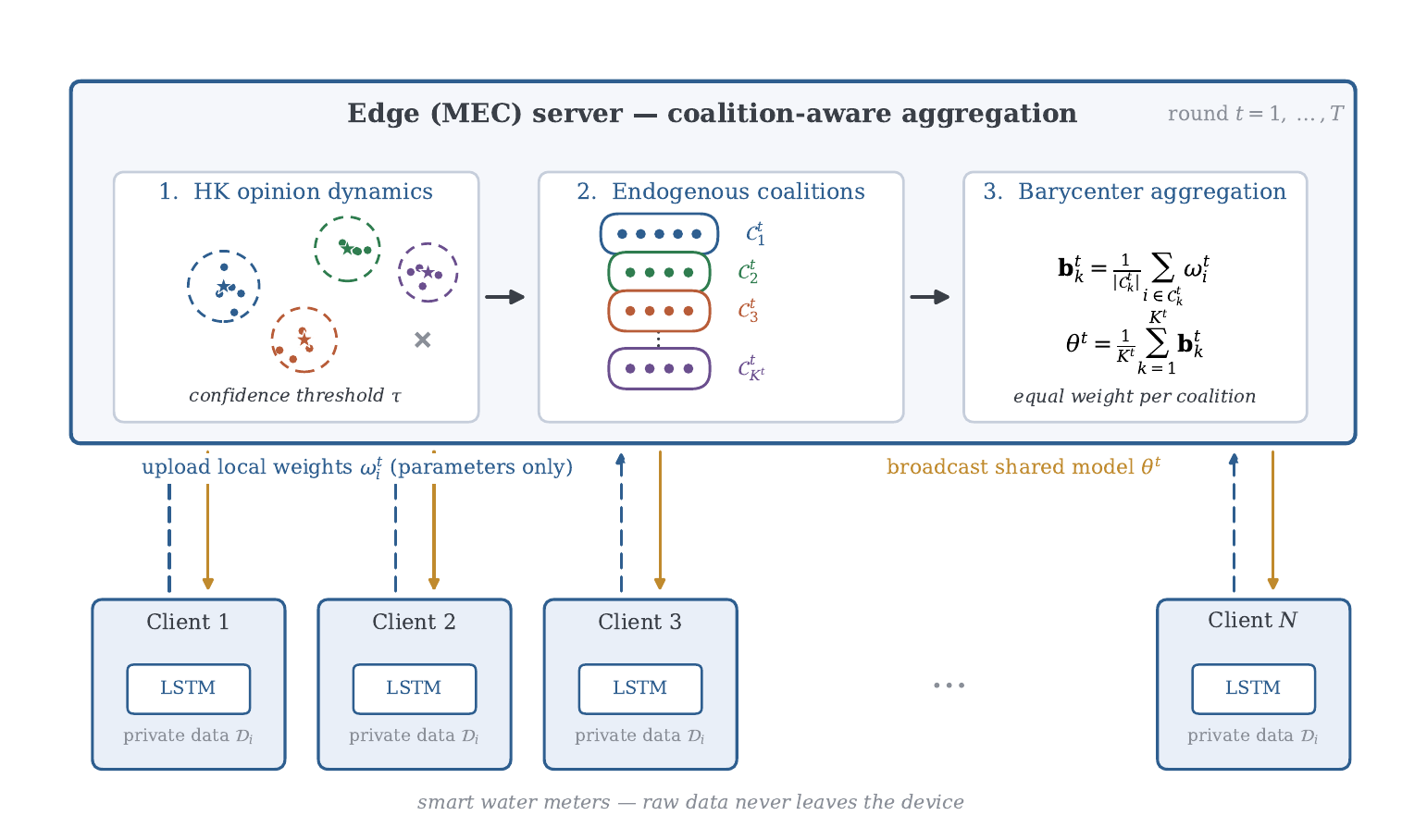}
    \caption{Proposed HK-based coalition-formation framework within one
    federated round $t$. Each client trains a local LSTM on its private data
    and uploads only the resulting weights $\bm{\omega}_i^{t}$. The edge (MEC)
    server then (1)~runs the bounded-confidence HK interaction on the local
    weights, (2)~extracts the endogenous coalitions $\{\mathcal{C}_k^{t}\}$,
    leaving atypical clients isolated as outliers, and (3)~aggregates the
    coalition barycenters $\mathbf{b}_k^{t}$, with equal weight per coalition,
    into the next shared model $\bm{\theta}^{t}$, which is broadcast back to
    the clients.}
    \label{fig:framework}
\end{figure*}

Following the problem statement of Section~\ref{ssec:pf_statement}, we treat each
client's local model as an opinion in parameter space and let opinions
interact only when they are mutually compatible. This is the HK
bounded-confidence model of opinion dynamics~\cite{rainer2002opinion}
applied to the local weights: an agent revises its position toward the average of
the agents inside its confidence neighborhood, so consensus forms within groups
of compatible agents while incompatible agents stay apart. Used on local models,
this yields an endogenous partition, where compatible clients converge to a
common position (a coalition) and atypical clients remain isolated, which is
exactly the adaptive behavior that a fixed-$K$ hard assignment cannot provide.
Figure~\ref{fig:framework} summarizes the resulting pipeline.

Concretely, at federated round $t$ the local models $\{\bm{\omega}_i^{t}\}_{i\in
\mathcal{S}^{t}}$ obtained after local training (Section~\ref{ssec:sm_protocol})
are used to initialize an inner HK iteration indexed by $s=0,1,\dots$:
\begin{equation}
    \bm{z}_i^{(0)} = \bm{\omega}_i^{t}, \qquad i\in\mathcal{S}^{t}.
\end{equation}
Given a confidence neighborhood $\mathcal{N}_i(\bm{z}^{(s)})\subseteq
\mathcal{S}^{t}$ (defined per variant below, and always with
$i\in\mathcal{N}_i$), each opinion is updated to the mean of its neighbors,
\begin{equation}
    \bm{z}_i^{(s+1)}
    = \frac{1}{\bigl|\mathcal{N}_i(\bm{z}^{(s)})\bigr|}
      \sum_{j\in\mathcal{N}_i(\bm{z}^{(s)})} \bm{z}_j^{(s)},
    \qquad i\in\mathcal{S}^{t}.
    \label{eq:hk_update}
\end{equation}
The iteration is run until the opinions stabilize, i.e.\ until
$\max_i\|\bm{z}_i^{(s+1)}-\bm{z}_i^{(s)}\|_2\le\delta$ for a small tolerance
$\delta>0$ (or a maximum number of inner steps $s_{\max}$ is reached); let
$\bm{z}_i^{\star}$ denote the resulting profile. The coalitions are the groups of
clients that have reached the same limiting opinion,
\begin{equation}
    i\sim j
    \;\Longleftrightarrow\;
    \bigl\|\bm{z}_i^{\star}-\bm{z}_j^{\star}\bigr\|_2 \le \delta ,
    \label{eq:coalition_classes}
\end{equation}
and the partition $\Pi^{t}=\{\mathcal{C}_1^{t},\dots,\mathcal{C}_{K^{t}}^{t}\}$
is the set of equivalence classes of $\sim$. The number of coalitions $K^{t}$ is
not prescribed: it is determined by the confidence parameter and by the
configuration of the local weights, and may vary across rounds. A single
application of \eqref{eq:hk_update} ($s_{\max}=1$) recovers the one-step
weight-averaging interaction; iterating to a fixed point is what produces
well-separated coalitions.

The three variants below differ only in how the confidence neighborhood
$\mathcal{N}_i$ is defined, i.e.\ in the geometry used to decide which opinions
are mutually compatible, building directly on the two geometries motivated in
Section~\ref{ssec:pf_similarity}.

\subsection{Euclidean HK interaction}
\label{ssec:fw_euclid}

The first variant measures compatibility by the Euclidean distance
\eqref{eq:deuclid} and admits as neighbors all opinions within a confidence ball
of radius $\varepsilon>0$:
\begin{equation}
    \mathcal{N}_i(\bm{z}^{(s)})
    = \Bigl\{\, j\in\mathcal{S}^{t} \;:\;
      \bigl\|\bm{z}_j^{(s)}-\bm{z}_i^{(s)}\bigr\|_2 \le \varepsilon \Bigr\}.
    \label{eq:nbr_euclid}
\end{equation}
This is the classical HK ball: clients whose models are close in absolute terms
reinforce one another. The radius $\varepsilon$ controls granularity (small
$\varepsilon$ yields many tight coalitions, large $\varepsilon$ merges them), and
its appropriate scale depends on the magnitude of the weights, which is the main
limitation addressed by the cosine variant next.\\
\\
In our implementation, the Euclidean variant operates on normalized
weight vectors: each opinion is first rescaled to unit norm,
$\hat{\bm{z}}_i^{(s)} = \bm{z}_i^{(s)}/\bigl\|\bm{z}_i^{(s)}\bigr\|_2$, before
evaluating \eqref{eq:nbr_euclid}. This removes the dependence of the distance
scale on the weight magnitudes: distances between unit vectors are bounded,
$\bigl\|\hat{\bm{z}}_i-\hat{\bm{z}}_j\bigr\|_2\in[0,2]$, and satisfy
$\bigl\|\hat{\bm{z}}_i-\hat{\bm{z}}_j\bigr\|_2^{2}
= 2\bigl(1-\mathrm{cosim}(\bm{z}_i,\bm{z}_j)\bigr)$,
so a confidence radius has a consistent geometric meaning across rounds and
model dimensions. It also makes the common threshold grid
$\{0.5,0.75,0.9\}$ used in Section~\ref{sec:results} directly comparable
across the two geometries (with a slight abuse of notation, we denote the
Euclidean radius by $\tau$ as well): both criteria then operate on a bounded,
norm-independent scale and differ only in whether compatibility is expressed
as a distance ball or as an angular alignment. Note that the two parameters
act in opposite directions: smaller $\tau$ yields finer coalitions in the
Euclidean case, whereas larger $\tau$ does so in the cosine case.

\subsection{Cosine-similarity HK interaction}
\label{ssec:fw_cosine}

The second variant measures compatibility by the cosine similarity
\eqref{eq:cosim} and admits as neighbors the opinions whose direction is
sufficiently aligned, through a threshold $\tau\in[-1,1]$:
\begin{equation}
    \mathcal{N}_i(\bm{z}^{(s)})
    = \Bigl\{\, j\in\mathcal{S}^{t} \;:\;
      \mathrm{cosim}\bigl(\bm{z}_i^{(s)},\bm{z}_j^{(s)}\bigr) \geq \tau \Bigr\}.
    \label{eq:nbr_cosine}
\end{equation}
which is equivalent to; $1-\operatorname{cosim}\bigl(z_i^{(s)},z_j^{(s)}\bigr) \leq 1-\tau$, perfectly aligned opinions ($\operatorname{cosim}=1$) are always neighbors, and $i \in \mathcal{N}_i$ holds automatically since
$\operatorname{cosim}\bigl(z_i^{(s)},z_i^{(s)}\bigr)=1$.\\
Because cosine similarity is invariant to the norm of the weight vectors, this
neighborhood groups clients with aligned learning directions even when their
models differ in magnitude (e.g.\ from quantity skew or unequal local progress),
a situation in which the Euclidean ball \eqref{eq:nbr_euclid} would wrongly
separate them. This makes the cosine variant better suited to high-dimensional
neural weights, where direction is often more informative than absolute distance.

\subsection{Asymmetric cosine-confidence model}
\label{ssec:fw_asym}

In practice the pairwise cosine similarities among local models tend to
concentrate near $1$, since the weights are correlated across clients. A single
threshold $\tau$ then offers limited control in that concentrated regime. We
therefore generalize \eqref{eq:nbr_cosine} to a two-sided confidence bound with
independent lower and upper bounds $\varepsilon_\ell,\varepsilon_r$ with
$-1\le\varepsilon_\ell\le\varepsilon_r\le 1$:
\begin{equation}
    \mathcal{N}_i(\bm{z}^{(s)})
    = \Bigl\{\, j\in\mathcal{S}^{t} \;:\;
      \varepsilon_\ell \le
      \mathrm{cosim}\bigl(\bm{z}_i^{(s)},\bm{z}_j^{(s)}\bigr)
      \le \varepsilon_r \Bigr\}.
    \label{eq:nbr_asym}
\end{equation}
The lower bound $\varepsilon_\ell$ excludes weakly aligned (dissimilar) clients,
as before, while the upper bound $\varepsilon_r$ allows excluding near-collinear
models when desired; the one-sided threshold variant \eqref{eq:nbr_cosine} is
recovered by setting $\varepsilon_r=1$ (so that $\tau=\varepsilon_\ell$). The
asymmetric bound gives finer coalitions in the high-similarity regime by carving
the narrow interval where most pairwise similarities lie. In the experiments we
use the one-sided threshold form, i.e.\ $\varepsilon_r=1$, with
$\tau\in\{0.5,0.75,0.9\}$ (Section~\ref{sec:results}).

\subsection{Coalition aggregation algorithm}
\label{ssec:fw_algo}

Once the partition $\Pi^{t}$ is obtained from the converged HK profile
\eqref{eq:coalition_classes}, aggregation is performed at the coalition level. For
each coalition $\mathcal{C}_k^{t}$ we compute its barycenter from the
original local models (not the smoothed opinions), preserving the
information learned locally,
\begin{equation}
    \mathbf{b}_k^{t}
    = \frac{1}{|\mathcal{C}_k^{t}|}
      \sum_{i\in\mathcal{C}_k^{t}} \bm{\omega}_i^{t},
    \label{eq:barycenter}
\end{equation}
and form the next shared model as the average of the coalition barycenters,
\begin{equation}
    \bm{\theta}^{t}
    = \frac{1}{K^{t}} \sum_{k=1}^{K^{t}} \mathbf{b}_k^{t}.
    \label{eq:coalition_agg}
\end{equation}
Aggregating barycenters rather than individual models, together with weighting
coalitions equally in \eqref{eq:coalition_agg}, limits the influence of any single
heterogeneous or outlier client on $\bm{\theta}^{t}$: an isolated client forms a
singleton coalition and contributes at most $1/K^{t}$, instead of the potentially
larger sample-weighted share it would receive under FedAvg \eqref{eq:fedavg}.
When all clients fall into a single coalition, \eqref{eq:coalition_agg} reduces to
a plain (uniform) average, recovering standard aggregation as a special case. The
overall procedure is summarized in Algorithm~\ref{alg:hk_coalition}.

\begin{algorithm}[t]
\caption{FedHK: Hegselmann–Krause Coalition Formation for Heterogeneous FL}
\label{alg:hk_coalition}
\begin{algorithmic}[1]
\STATE Initialize global model $\bm{\theta}^{0}$; set confidence parameter
       ($\varepsilon$, $\tau$, or $(\varepsilon_\ell,\varepsilon_r)$),
       tolerance $\delta$, inner cap $s_{\max}$.
\FOR{each round $t=1,2,\dots,T$}
    \STATE Server selects active set $\mathcal{S}^{t}\subseteq\mathcal{N}$ and
           broadcasts $\bm{\theta}^{t-1}$.
    \FOR{each client $i\in\mathcal{S}^{t}$ \textbf{in parallel}}
        \STATE $\bm{\omega}_i^{t}\leftarrow
               \textsc{ClientUpdate}(i,\bm{\theta}^{t-1})$
               \COMMENT{$E$ local SGD epochs, Eq.~\eqref{eq:local_sgd}}
    \ENDFOR
    \STATE \textbf{HK interaction:} $\bm{z}_i^{(0)}\leftarrow\bm{\omega}_i^{t}$,
           $\forall i\in\mathcal{S}^{t}$; \; $s\leftarrow 0$.
    \REPEAT
        \FOR{each client $i\in\mathcal{S}^{t}$}
            \STATE Form $\mathcal{N}_i(\bm{z}^{(s)})$ via
                   Eq.~\eqref{eq:nbr_euclid}, \eqref{eq:nbr_cosine},
                   or \eqref{eq:nbr_asym}.
            \STATE $\bm{z}_i^{(s+1)}\leftarrow
                   \frac{1}{|\mathcal{N}_i(\bm{z}^{(s)})|}
                   \sum_{j\in\mathcal{N}_i(\bm{z}^{(s)})}\bm{z}_j^{(s)}$
                   \COMMENT{Eq.~\eqref{eq:hk_update}}
        \ENDFOR
        \STATE $s\leftarrow s+1$.
    \UNTIL{$\max_i\|\bm{z}_i^{(s)}-\bm{z}_i^{(s-1)}\|_2\le\delta$
           \textbf{ or } $s\ge s_{\max}$}
    \STATE Extract coalitions $\Pi^{t}=\{\mathcal{C}_1^{t},\dots,
           \mathcal{C}_{K^{t}}^{t}\}$ from $\bm{z}^{\star}$ via
           Eq.~\eqref{eq:coalition_classes}.
    \FOR{each coalition $\mathcal{C}_k^{t}$}
        \STATE $\mathbf{b}_k^{t}\leftarrow
               \frac{1}{|\mathcal{C}_k^{t}|}
               \sum_{i\in\mathcal{C}_k^{t}}\bm{\omega}_i^{t}$
               \COMMENT{Eq.~\eqref{eq:barycenter}}
    \ENDFOR
    \STATE $\bm{\theta}^{t}\leftarrow
           \frac{1}{K^{t}}\sum_{k=1}^{K^{t}}\mathbf{b}_k^{t}$
           \COMMENT{Eq.~\eqref{eq:coalition_agg}}
\ENDFOR
\STATE \textbf{return} $\bm{\theta}^{T}$
\end{algorithmic}
\end{algorithm}

In terms of complexity, beyond standard FedAvg, each round adds the HK
interaction. One inner step
computes pairwise compatibilities over the active set, costing
$O(S^{2}d)$ for $S=|\mathcal{S}^{t}|$ participating clients and parameter
dimension $d$; with at most $s_{\max}$ inner steps the per-round overhead is
$O(s_{\max}S^{2}d)$. Since the HK iteration and coalition extraction run on the
server over already-received models, no extra client computation or communication
is incurred relative to FedAvg, so the communication cost per round is unchanged.

\section{Experimental Setup}
\label{sec:setup}

\subsection{Data preprocessing}
\label{ssec:exp_data}

We evaluate the framework on the Smart Water Meter Consumption Time Series
dataset from the city of Alicante, Spain~\cite{alicante_dataset}, which provides
hourly consumption measurements for $N_{\text{total}}=1{,}099$ residential
clients. Because residential consumption profiles vary widely across households,
the dataset is a suitable testbed for heterogeneity-aware FL. We apply a
three-stage preprocessing pipeline: active-client filtering, federated cohort
sampling, and temporal aggregation.

For the first stage, let $C=\{1,\dots,N_{\text{total}}\}$ be the full client set and $\bar{x}_i$ the
mean hourly consumption of client $i$ over the observation period, with global
mean $\mu_C=\frac{1}{|C|}\sum_{i\in C}\bar{x}_i$. To retain clients with
consistently measurable consumption, we keep those whose mean exceeds the global
average by at least $60\%$,
\begin{equation}
    C_{\text{active}}=\bigl\{\, i\in C \;:\; \bar{x}_i \ge 1.6\,\mu_C \,\bigr\},
    \label{eq:active_filter}
\end{equation}
which yields $C_{\text{active}} = 188$ clients and excludes
predominantly zero or minimal-consumption clients.

Of the active clients, $184$ with at least two years of recorded data form
the federated pool $C_{\text{FL}}$. At each communication round the server
samples, uniformly without replacement, an active subset of $S=50$ clients from
this pool (Section~\ref{ssec:sm_protocol}), which keeps the experiments
computationally tractable and mitigates selection bias.

Finally, hourly residential consumption is highly sparse and zero-inflated. For
each client
$i\in C_{\text{FL}}$ with hourly series
$\mathbf{z}^{(i)}=(z_1^{(i)},\dots,z_T^{(i)})$, we aggregate consumption into
non-overlapping $k=24$-hour windows,
\begin{equation}
    y_j^{(i,k)}=\sum_{t=(j-1)k+1}^{\min(j\cdot k,\,T)} z_t^{(i)},
    \qquad j=1,\dots,\lfloor T/k\rfloor,
    \label{eq:temporal_agg}
\end{equation}
which markedly reduces sparsity (Table~\ref{tab:dataset_stats}). The aggregated
daily series are then normalized and converted into the supervised one-step-ahead
form of Section~\ref{ssec:sm_data} using sliding look-back windows of length
$L=12$ time steps, and split chronologically into $80\%$ training and $20\%$
test partitions.

\begin{table}[!htbp]
\centering
\caption{Dataset characteristics before and after temporal aggregation.}
\label{tab:dataset_stats}
\begin{tabular}{lccc}
\toprule
\textbf{Resolution} & \textbf{Samples/year} & \textbf{Zero rate (\%)} & \textbf{Mean consumption} \\
\midrule
Hourly (original) & 8{,}760 & 77.2 & 5.38 \\
24-hour (aggregated) & 365 & 12.2 & 139.12 \\
\bottomrule
\end{tabular}
\end{table}
\subsection{Baselines}
\label{ssec:exp_baselines}

We compare the two coalition variants of Section~\ref{sec:framework} against two
heterogeneity-oriented FL baselines:
\begin{itemize}
    \item \textbf{FedAvg with Euclidean coalitions}: the proposed framework with
    the Euclidean confidence ball \eqref{eq:nbr_euclid}.
    \item \textbf{FedAvg with cosine coalitions}: the proposed framework with the
    cosine-similarity threshold \eqref{eq:nbr_cosine}.
    \item \textbf{FedProx}~\cite{li2020federated}: FedAvg with a proximal term
    that limits client drift under heterogeneity.
    \item \textbf{Per-FedAvg}~\cite{reguieg2023comparative}: a
    meta-learning-based personalized FL method that adapts the shared model to
    each client.
\end{itemize}
All methods share the same LSTM architecture, optimizer and round budget so that
differences are attributable to the aggregation/personalization mechanism alone.

\subsection{Implementation and hyperparameters}
\label{ssec:exp_impl}

All experiments were run on an ASUS TUF A15 (AMD Ryzen~7 6800H at 4.7~GHz, 16~GB
RAM, NVIDIA RTX~3070~Ti). The federated pool consists of the $184$ clients with
at least two years of recorded data; at each of the $T=20$ communication rounds
the server samples $S=50$ of them uniformly at random
(Section~\ref{ssec:sm_protocol}). Each selected client runs $E=5$ local epochs
of Adam ($\eta=10^{-3}$, batch size $32$, MSE loss) on its own series, split
chronologically into $80\%$ training and $20\%$ test data, with a look-back
window of $L=12$ time steps. The HK interaction operates on the post-training
local weights (Section~\ref{ssec:fw_hk}), unit-normalized in the Euclidean
case (Section~\ref{ssec:fw_euclid}), and is evaluated with both the
Euclidean and the cosine compatibility geometries, each with confidence
thresholds $\tau\in\{0.5,0.75,0.9\}$; the inner iteration stabilizes within
at most ten steps in all settings (Section~\ref{ssec:res_coalitions}). The full
hyperparameter configuration is summarized in Table~\ref{tab:hyperparams}.

\begin{table}[!htbp]
\centering
\caption{Hyperparameter configuration.}
\label{tab:hyperparams}
\begin{tabular}{lll}
\toprule
\textbf{Component} & \textbf{Hyperparameter} & \textbf{Value} \\
\midrule
\multirow{4}{*}{Data}
  & Aggregation window $k$                 & $24$\,h \\
  & Active-client threshold                & $1.6\,\mu_C$ \\
  & Look-back window length $L$            & $12$ time steps \\
  & Train/test split ratio                 & $80\%$ / $20\%$ \\
\midrule
\multirow{5}{*}{LSTM model}
  & Input size                             & $1$ \\
  & LSTM layers (units)                    & $2$ ($32$, $16$) \\
  & Dropout rate                           & $0.2$ \\
  & Read-out layers                        & Dense $16\!\to\!8\!\to\!1$, ReLU \\
  & Output activation                      & ReLU (non-negative predictions) \\
\midrule
\multirow{5}{*}{Local training}
  & Optimizer                              & Adam \\
  & Local learning rate $\eta$             & $1\times10^{-3}$ \\
  & Batch size                             & $32$ \\
  & Local epochs $E$                       & $5$ \\
  & Loss                                   & MSE \\
\midrule
\multirow{4}{*}{Federated protocol}
  & Total clients $N_{\text{total}}$       & $1{,}099$ \\
  & Client pool ($\ge 2$ years of data)    & $184$ \\
  & Clients per round $S$                  & $50$ (randomly sampled) \\
  & Communication rounds $T$               & $50$ \\
\midrule
\multirow{3}{*}{HK coalition}
  & Opinion vector                         & local weights $\bm{\omega}_i^{t}$ \\
  & Compatibility geometries               & Euclidean, cosine \\
  & Confidence thresholds $\tau$           & $\{0.5, 0.75, 0.9\}$ \\
\bottomrule
\end{tabular}
\end{table}

\subsection{Evaluation metrics}
\label{ssec:exp_metrics}

Forecasting accuracy is reported with the Mean Squared Error (MSE) and the
Mean Absolute Error (MAE),
\begin{equation}
    \mathrm{MSE}=\frac{1}{P}\sum_{i=1}^{P}\bigl(z_i-\hat{z}_i\bigr)^2,
    \qquad
    \mathrm{MAE}=\frac{1}{P}\sum_{i=1}^{P}\bigl|z_i-\hat{z}_i\bigr|,
    \label{eq:metrics}
\end{equation}
where $z_i$ and $\hat{z}_i$ are the actual and predicted values and $P$ is the
number of predictions. The MSE on the held-out test set, tracked as a function
of the communication round, is used for the convergence comparison across
methods, while the MAE, averaged over the client cohort, is used for the final
head-to-head accuracy comparison.

\section{Experimental Results and Discussion}
\label{sec:results}

\subsection{Coalition structure under HK dynamics}
\label{ssec:res_coalitions}

\begin{figure}[!htbp]
    \centering
    \subfloat[Euclidean, $\tau=0.5$]{%
        \includegraphics[width=0.48\textwidth]{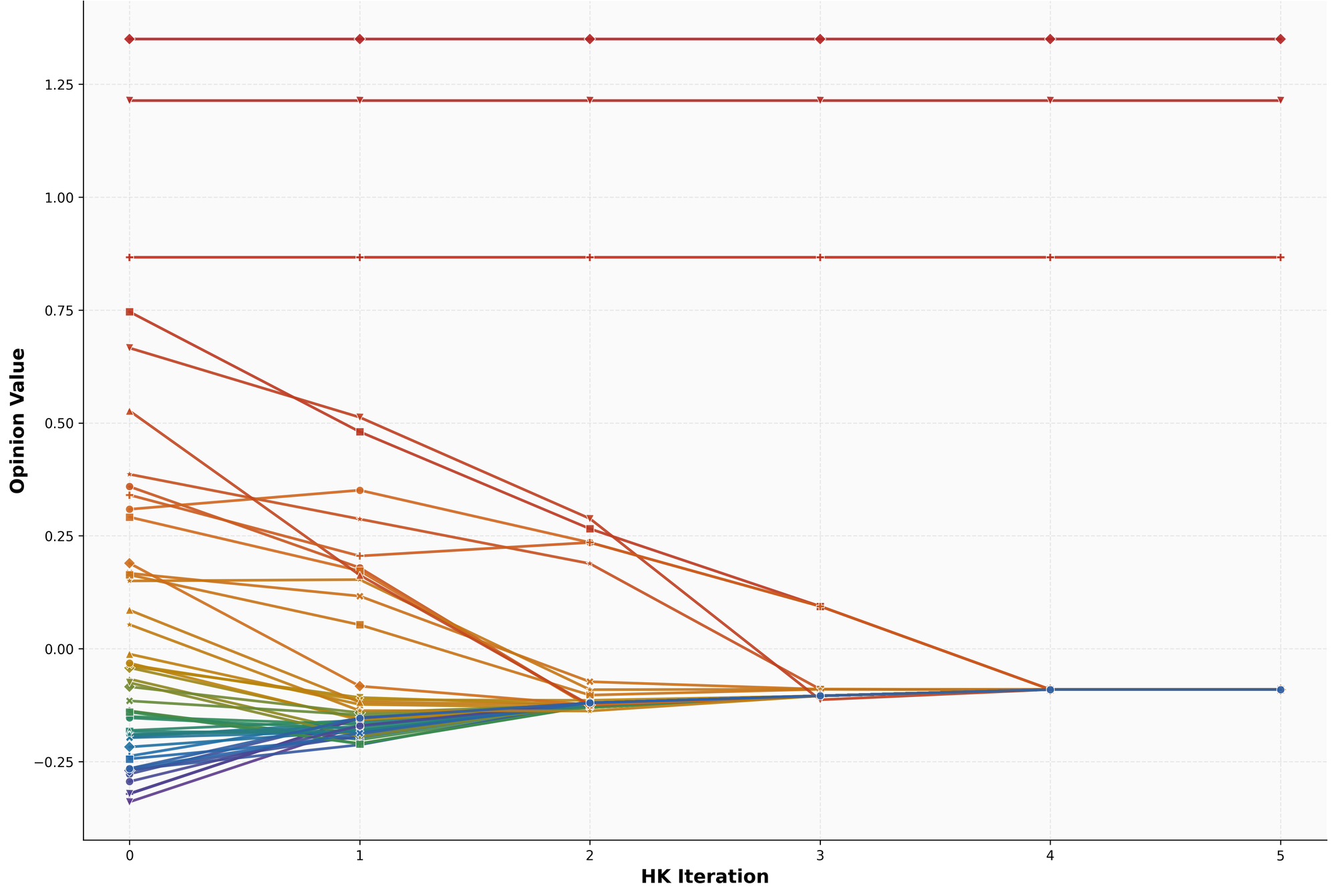}%
        \label{fig:hk_traj_euc_05}}\hfill
    \subfloat[Cosine, $\tau=0.5$]{%
        \includegraphics[width=0.48\textwidth]{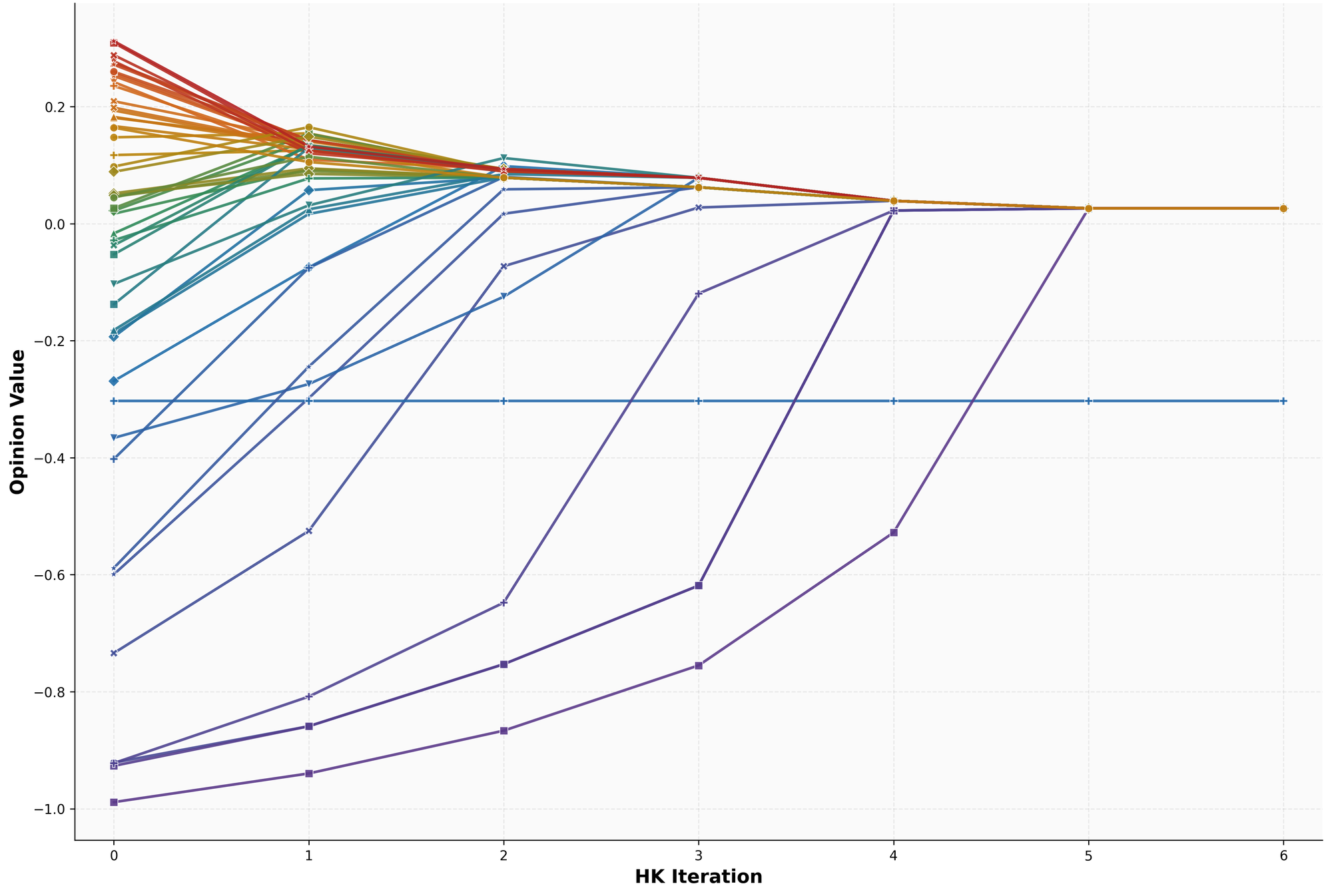}%
        \label{fig:hk_traj_cos_05}}\\
    \subfloat[Euclidean, $\tau=0.75$]{%
        \includegraphics[width=0.48\textwidth]{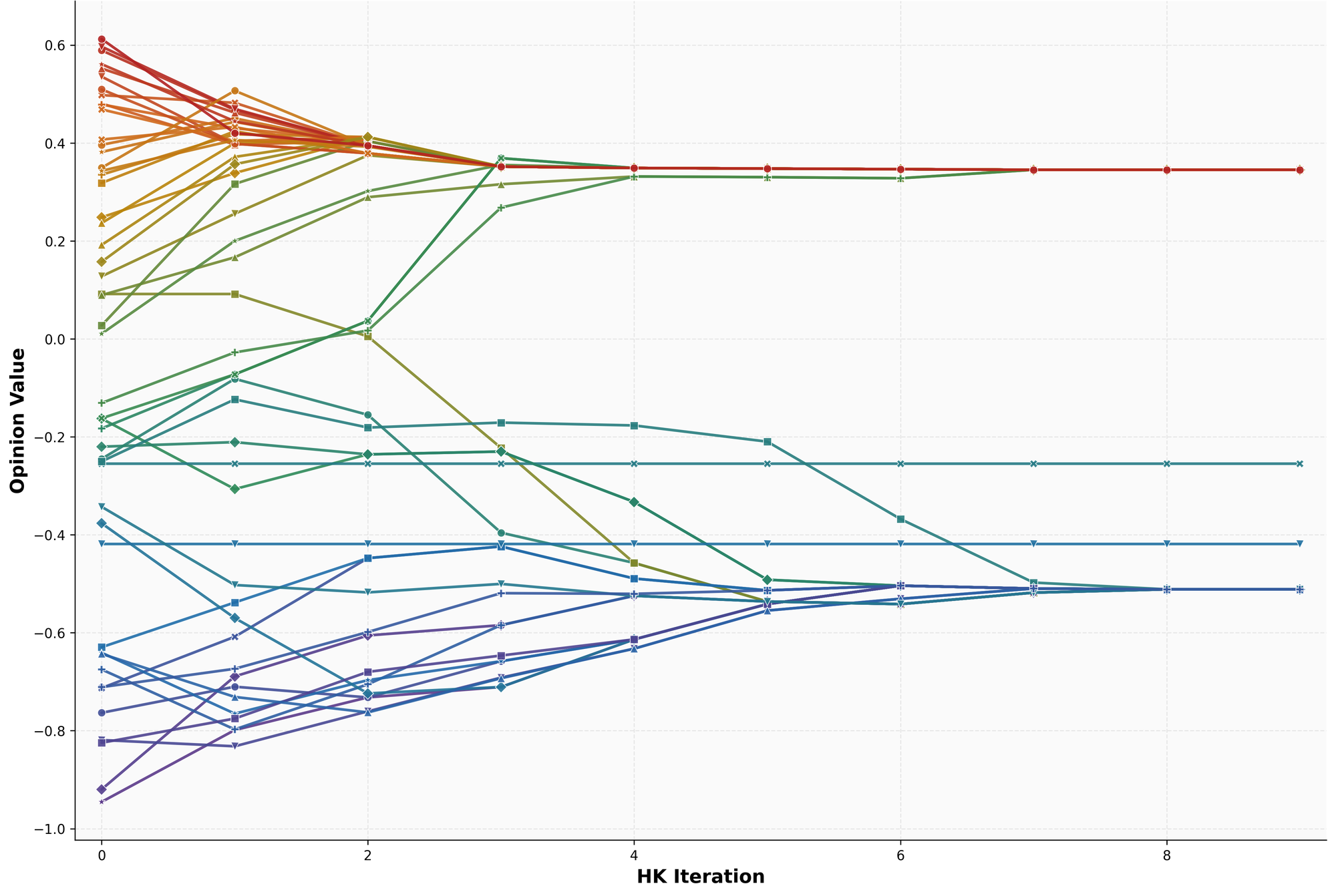}%
        \label{fig:hk_traj_euc_075}}\hfill
    \subfloat[Cosine, $\tau=0.75$]{%
        \includegraphics[width=0.48\textwidth]{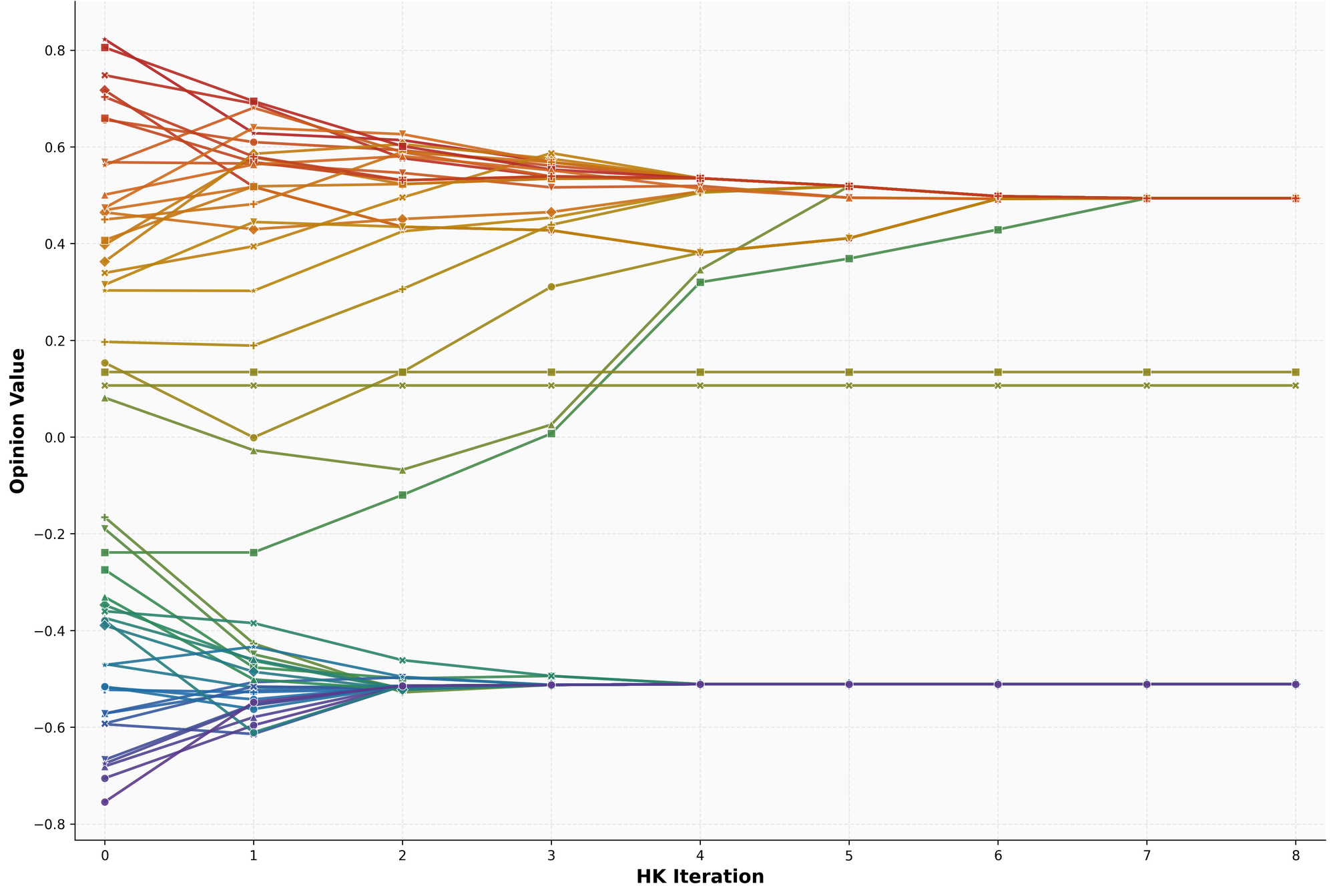}%
        \label{fig:hk_traj_cos_075}}\\
    \subfloat[Euclidean, $\tau=0.9$]{%
        \includegraphics[width=0.48\textwidth]{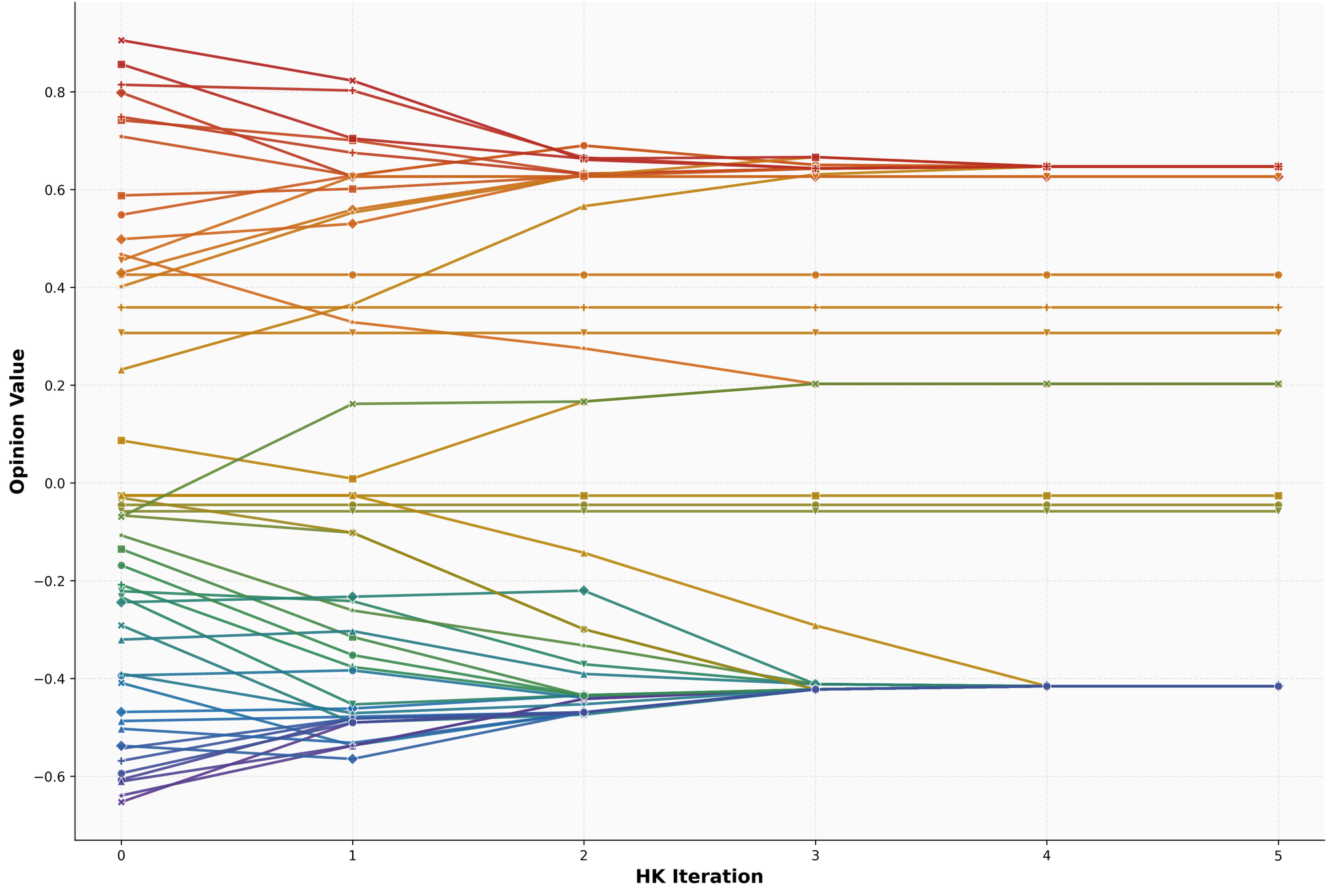}%
        \label{fig:hk_traj_euc_09}}\hfill
    \subfloat[Cosine, $\tau=0.9$]{%
        \includegraphics[width=0.48\textwidth]{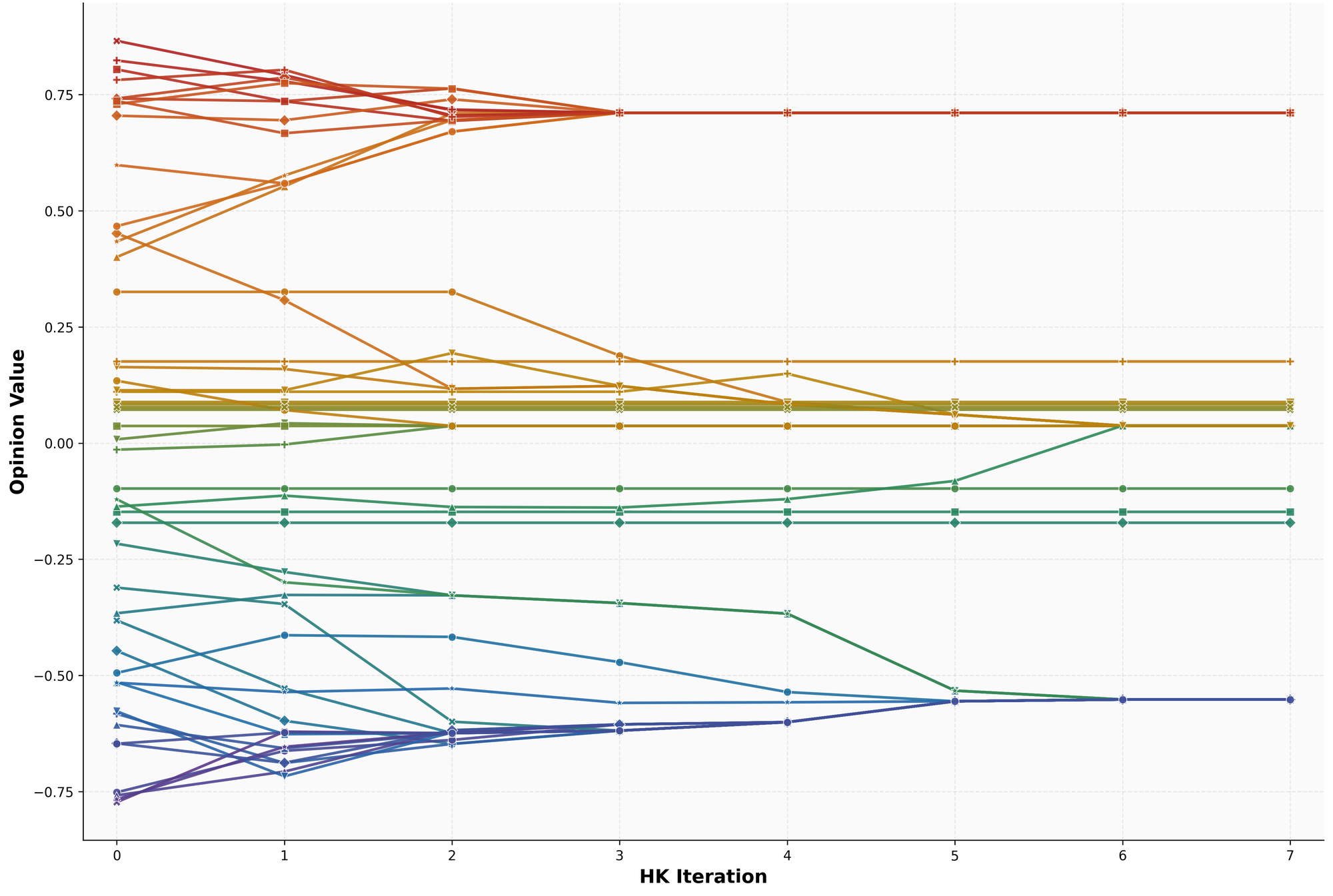}%
        \label{fig:hk_traj_cos_09}}
    \caption{HK opinion trajectories of the local weights under the Euclidean
    (left) and cosine (right) interactions for confidence thresholds
    $\tau\in\{0.5,0.75,0.9\}$. The permissive threshold (a)--(b) yields a
    single large coalition with a few isolated outliers; the intermediate
    threshold (c)--(d) yields two well-separated macro-coalitions; the
    tightest threshold (e)--(f) yields a finer partition with more coalitions
    and singletons. All settings converge within at most ten inner
    iterations.}
    \label{fig:hk_traj}
\end{figure}

Figure~\ref{fig:hk_traj} shows the HK opinion trajectories of the local
weights for the Euclidean (left) and cosine (right) interactions, with
thresholds $\tau\in\{0.5,0.75,0.9\}$. In all six settings the opinions
stabilize within at most ten inner iterations, which supports the small
inner cap $s_{\max}$ used in the complexity analysis
(Section~\ref{ssec:fw_algo}). The threshold sets how fine the partition is,
in the same way for both geometries. With $\tau=0.5$
(Figures~\ref{fig:hk_traj_euc_05} and~\ref{fig:hk_traj_cos_05}) almost all
clients merge into one large coalition, while a few atypical clients never
interact and stay isolated. With $\tau=0.75$
(Figures~\ref{fig:hk_traj_euc_075} and~\ref{fig:hk_traj_cos_075}) the
population splits into two well-separated coalitions, again with a few
isolated clients. With $\tau=0.9$
(Figures~\ref{fig:hk_traj_euc_09} and~\ref{fig:hk_traj_cos_09}) several
smaller coalitions appear, along with more singletons. The isolated clients
are households with unusual consumption (e.g.\ very large occupancy or
non-residential use). The bounded-confidence rule treats them as outliers
instead of forcing them into a coalition, as a fixed-$K$ scheme would. This
is the intended behavior (Section~\ref{ssec:fw_hk}): the number of
coalitions is endogenous and is controlled by a single parameter.

\subsection{Convergence comparison}
\label{ssec:res_convergence}

\begin{figure}[!htbp]
    \centering
    \includegraphics[width=\textwidth]{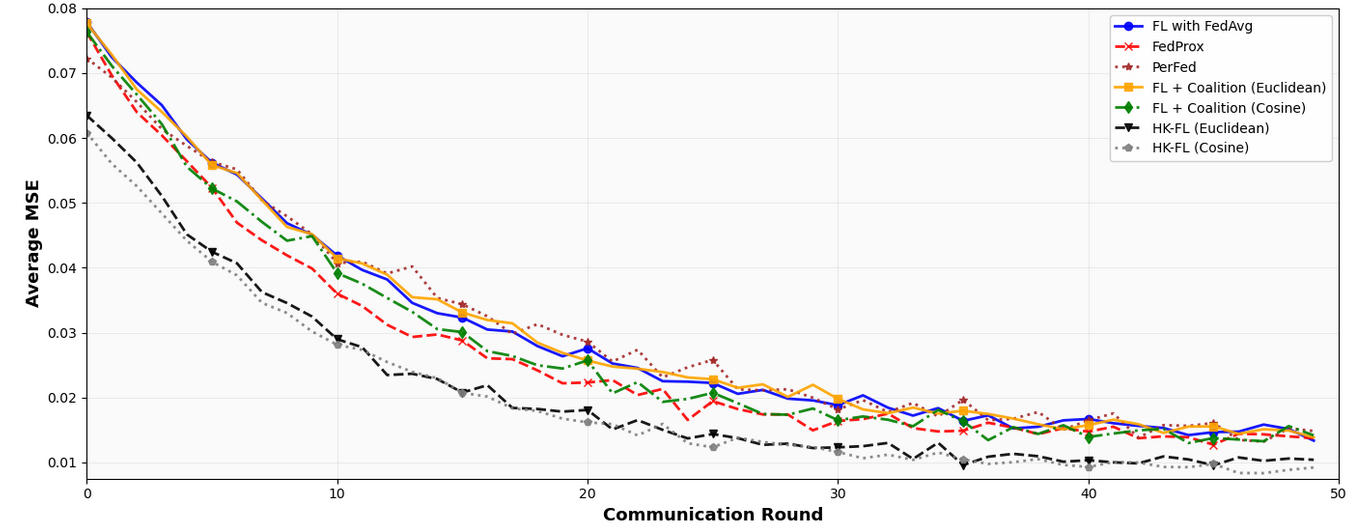}
    \caption{Average held-out MSE per communication round for FedAvg,
    Per-FedAvg, FedProx, the fixed-$K$ coalition scheme
    of~\cite{el2024efficient} (FL~+~Coalition, Euclidean and cosine), and the
    proposed HK-based variants (HK-FL, Euclidean and cosine).}
    \label{fig:mse_comp}
\end{figure}

\begin{figure}[!htbp]
    \centering
    \includegraphics[width=\textwidth]{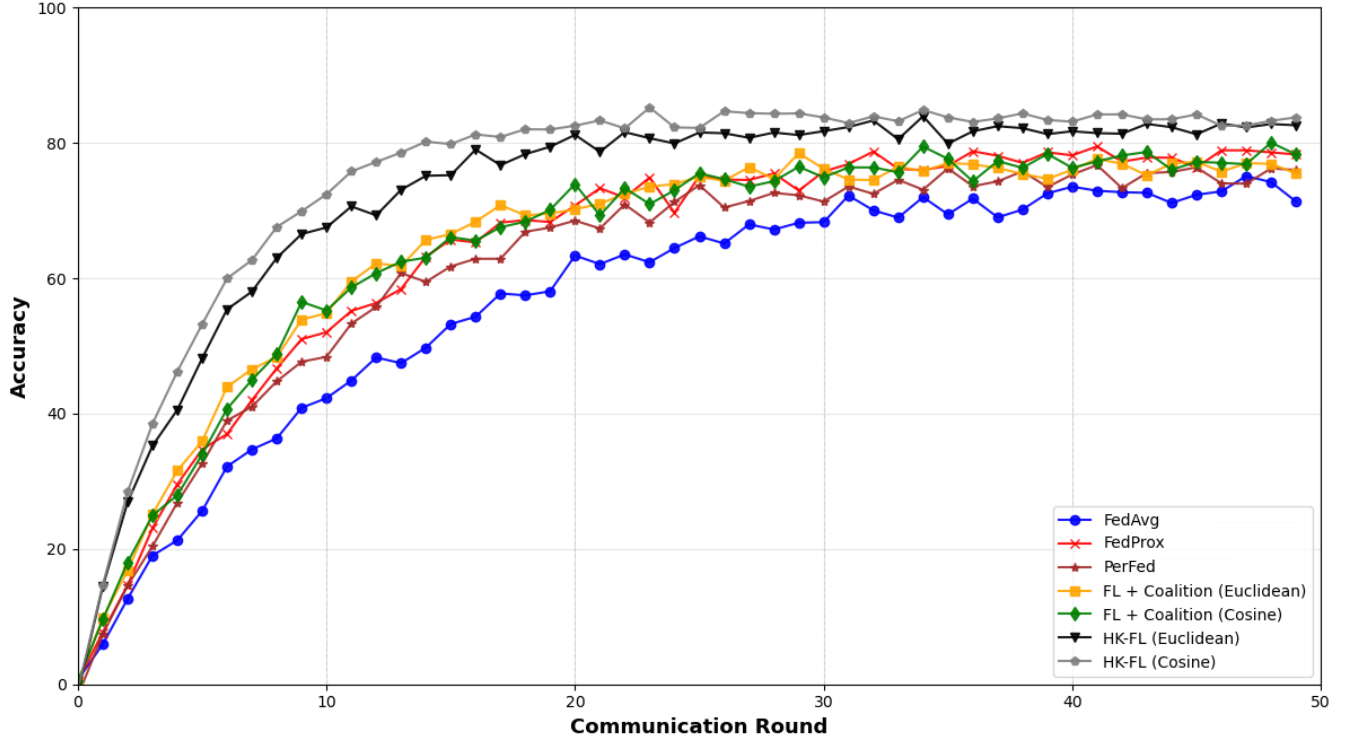}
    \caption{Global-model accuracy per communication round for the same
    methods as in Figure~\ref{fig:mse_comp}. The proposed HK-FL variants reach
    the highest accuracy, with the cosine variant slightly above the Euclidean
    one.}
    \label{fig:acc_comp}
\end{figure}

Figure~\ref{fig:mse_comp} compares the average held-out MSE of the methods
across the communication rounds: FedAvg, Per-FedAvg, FedProx, the fixed-$K$
coalition scheme of~\cite{el2024efficient} (FL~+~Coalition, Euclidean and
cosine), and the two proposed HK variants (HK-FL). Three points stand out.
First, the HK-FL variants are the best throughout training: they start
lower, decay faster, and reach the lowest final MSE (about
$0.010$--$0.011$, versus $0.014$--$0.015$ for all baselines;
Table~\ref{tab:method_comparison}). Dissimilar clients stop averaging into
one another early, so useful shared models appear sooner. Second, the
fixed-$K$ baselines track FedAvg closely: the gain of HK-FL comes from the
endogenous partition, not from coalition averaging itself. Third, FedProx
improves slightly on FedAvg, while Per-FedAvg is noisier and does not; both
end well above HK-FL, so neither regularization nor personalization alone
handles the heterogeneity. Figure~\ref{fig:acc_comp} shows the accuracy per
round and confirms this ordering: HK-FL plateaus at $83$--$85\%$, the other
baselines at $75$--$79\%$, and FedAvg lowest, with the cosine variant
slightly above the Euclidean one.

\subsection{Comparison across methods}
\label{ssec:res_quantitative}

\begin{figure}[!htbp]
    \centering
    \includegraphics[width=\textwidth]{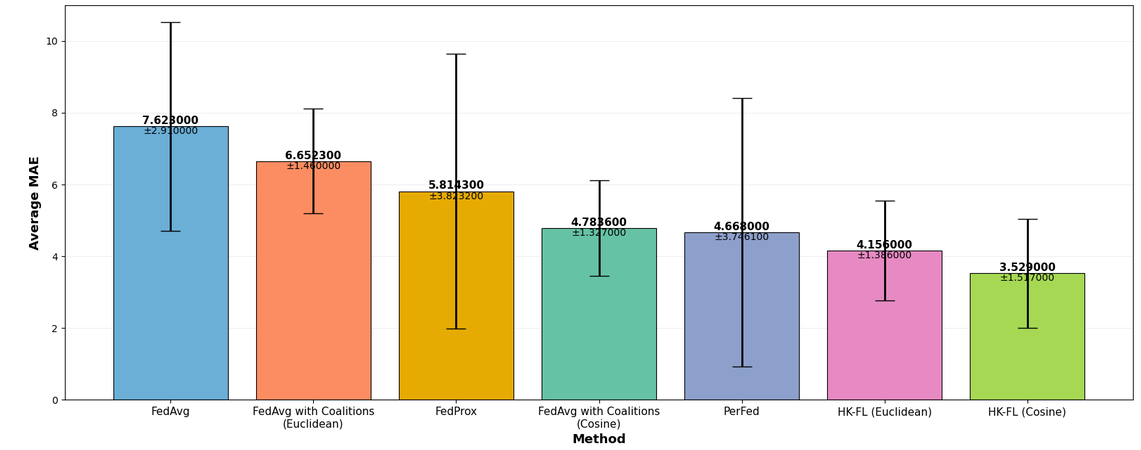}
    \caption{Average MAE over the evaluated clients (error bars:
    $\pm$ one standard deviation across clients) for the compared methods.}
    \label{fig:mae_bar}
\end{figure}

Figure~\ref{fig:mae_bar} and Table~\ref{tab:method_comparison} report the
final Mean Absolute Error (MAE), averaged over the evaluated clients
(mean~$\pm$~std). The HK-FL variants obtain the two lowest errors: the
cosine variant cuts the average MAE by about $54\%$ versus FedAvg ($3.53$
versus $7.62$), $39\%$ versus FedProx, and $24\%$ versus Per-FedAvg.
Replacing the fixed-$K$ partition with the endogenous HK partition lowers
the MAE from $6.65$ to $4.16$ (Euclidean) and from $4.78$ to $3.53$
(cosine), which isolates the contribution of the bounded-confidence
mechanism. Two more patterns are worth noting. First, cosine beats Euclidean
within each scheme, supporting the argument of
Section~\ref{ssec:pf_similarity} that the direction of high-dimensional
weights is more informative than their distance. Second, the HK variants
have small cross-client spreads ($\pm1.39$ and $\pm1.52$, versus $\pm3.82$
for FedProx and $\pm3.75$ for Per-FedAvg): the gain is shared across the
cohort, which we attribute to outliers being isolated instead of
contaminating the shared model.

\begin{table}[!htbp]
\centering
\caption{Final forecasting error: MAE (mean~$\pm$~std across clients) and
final held-out MSE (approximate values read from
Figure~\ref{fig:mse_comp}); lower is better.}
\label{tab:method_comparison}
\begin{tabular}{lcc}
\toprule
\textbf{Method} & \textbf{MAE} & \textbf{MSE} ($\approx$) \\
\midrule
FedAvg~\cite{mcmahan2017communication}                & $7.62\pm2.91$ & $0.015$ \\
FedAvg with Euclidean coalitions~\cite{el2024efficient} & $6.65\pm1.46$ & $0.015$ \\
FedProx~\cite{li2020federated}                        & $5.81\pm3.82$ & $0.014$ \\
FedAvg with cosine coalitions~\cite{el2024efficient}  & $4.78\pm1.33$ & $0.014$ \\
Per-FedAvg~\cite{reguieg2023comparative}              & $4.67\pm3.75$ & $0.014$ \\
HK-FL, Euclidean (proposed)                           & $4.16\pm1.39$ & $0.011$ \\
HK-FL, cosine (proposed)                              & $\bm{3.53\pm1.52}$ & $\bm{0.010}$ \\
\bottomrule
\end{tabular}
\end{table}

\subsection{Client-level forecasting accuracy}
\label{ssec:res_clientlevel}

\begin{figure}[H]
% \begin{figure}[!htbp]
    \centering
    \subfloat[Client user~97]{%
        \includegraphics[width=\textwidth]{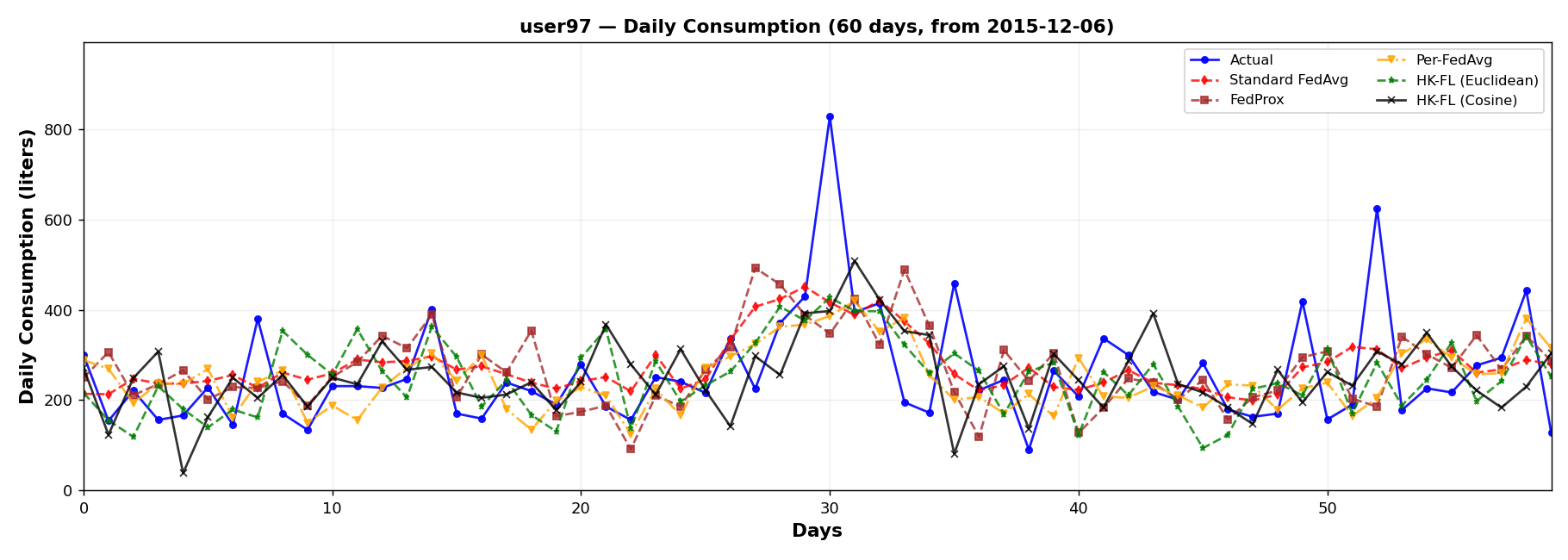}%
        \label{fig:client_daily_97}}\\
    \subfloat[Client user~108]{%
        \includegraphics[width=\textwidth]{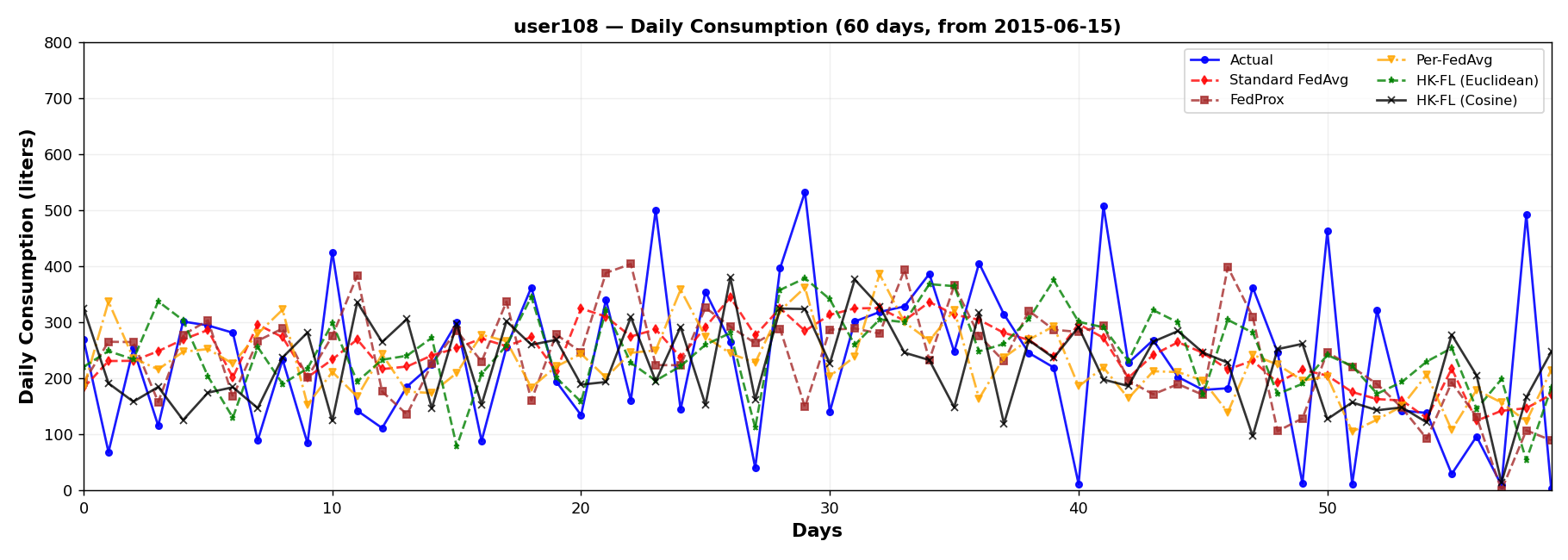}%
        \label{fig:client_daily_108}}\\
    \subfloat[Client user~715]{%
        \includegraphics[width=\textwidth]{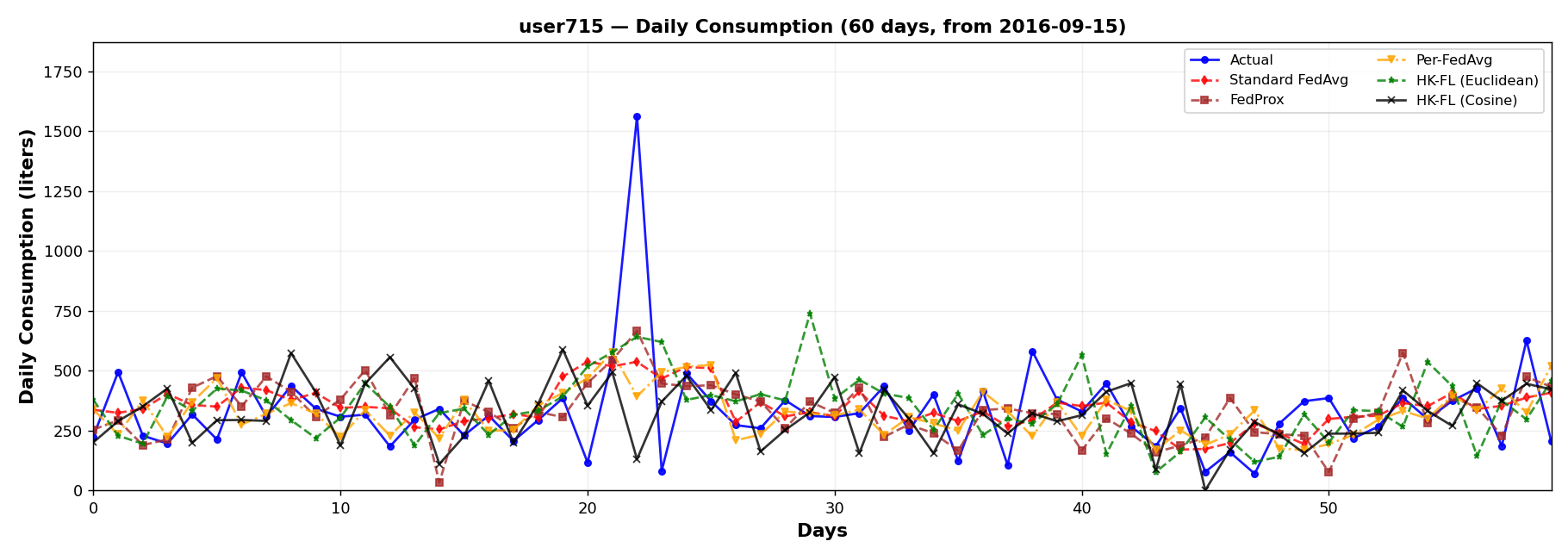}%
        \label{fig:client_daily_715}}
    \caption{Actual versus predicted daily water consumption over 60 days for
    three representative clients, comparing FedAvg, FedProx, Per-FedAvg, and
    the proposed HK-FL variants.}
    \label{fig:client_daily}
\end{figure}

Figure~\ref{fig:client_daily} shows 60 days of actual and predicted daily
consumption for three representative clients. All methods capture the
routine consumption level. The single-model baselines (FedAvg, FedProx)
smooth toward a common mean, while the HK-FL forecasts follow the
day-to-day changes of each household more closely, such as the
high-consumption episode around days 25--35 of user~97. This matches the
design: averaging only within compatible coalitions preserves
client-specific structure that a global average smooths out. Extreme
one-day spikes (about $800$ liters for user~97 and $1{,}500$ liters for
user~715) are missed by every method; such events are hard to predict from
past consumption alone, and this limitation is not specific to coalition
formation.\\
 
In summary, HK coalition formation on the local weights yields stable,
interpretable partitions controlled by a single threshold
(Section~\ref{ssec:res_coalitions}), and outperforms FedAvg, FedProx,
Per-FedAvg, and the fixed-$K$ scheme in convergence, final loss, accuracy,
and MAE (Sections~\ref{ssec:res_convergence}--\ref{ssec:res_quantitative}),
while isolating outliers instead of letting them distort the shared model.
The fixed-$K$ comparison shows that the endogenous partition drives the
gain, and the consistent edge of the cosine geometry supports
 direction-based compatibility for high-dimensional weights.
% \subsection{Discussion and limitations}
% \label{ssec:res_discussion}

% The main
% limitations are a single dataset and configuration, some undocumented
% hyperparameters (Table~\ref{tab:hyperparams}), and the missed extreme
% events (Section~\ref{ssec:res_clientlevel}); multi-seed statistics would
% strengthen the claims. We report these as observations on one dataset, not
% as claims of general superiority.

\section{Conclusion and Perspectives}
\label{sec:conclusion}

We addressed statistical heterogeneity in FL for heterogeneous
IoT systems by forming client coalitions directly in the local-weight space and
aggregating at the coalition level. Casting coalition formation as an
HK bounded-confidence interaction on the local weights yields a
partition whose number and membership are endogenous, and which isolates atypical
clients as outliers rather than forcing them into a coalition. We instantiated the
interaction with three compatibility geometries (a Euclidean confidence ball, a
cosine-similarity threshold, and an asymmetric cosine-confidence bound) and gave a
coalition aggregation algorithm that adds no client computation or communication
over FedAvg.

Instantiated for short-term water-consumption forecasting with local LSTM models
on the Alicante smart-meter dataset, the framework produced stable, interpretable
coalition structures and outperformed FedAvg, FedProx, Per-FedAvg, and a
fixed-$K$ weight-driven coalition baseline: the HK variants converged faster to
the lowest held-out MSE and highest accuracy, and the cosine variant reduced the
cohort-averaged MAE by about $54\%$ relative to FedAvg while explicitly
surfacing atypical clients as outliers. We report these as empirical
observations on a single dataset and do not claim general superiority of any
single method.

Several directions remain. On the methodological side, promising extensions
include a sensitivity analysis of the confidence parameters and an adaptive
schedule for them, a theoretical study of the convergence of coalition-level
aggregation under the bounded-heterogeneity condition, the integration of secure
aggregation or differential privacy on the uploaded updates, and the use of
coalition-specific models for personalization rather than a single global model.

\backmatter

\bmhead{Acknowledgements}
This work is supported by the Alkhawarizmi AI Project (grant number: Alkhawarizmi/2020/34).

\section*{Declarations}
\begin{description}
  \item[\textbf{Ethics declaration:}] Not applicable.
  \item[\textbf{Conflict of interest:}] The authors declare that they have no conflict of interest.

\end{description}

\bibliography{ref_journal}

\end{document}